\documentclass[twoside,11pt]{article}
\usepackage[abbrvbib, preprint]{jmlr2e}
\usepackage{blindtext}
\usepackage{enumitem}

\usepackage{amsmath}
\usepackage{makecell}
\usepackage{tcolorbox}
\usepackage{booktabs}
\usepackage{threeparttable}
\usepackage{multirow}

\usepackage{graphicx}
\usepackage{subcaption}
\usepackage{pifont}

\usepackage{listings}

\definecolor{darkgreen}{rgb}{0,0.6,0}
\definecolor{gray}{rgb}{0.5,0.5,0.5}
\definecolor{mauve}{rgb}{0.58,0,0.82}

\definecolor{logocolor}{RGB}{30, 0, 178}

\usepackage{jmlr2e}
\usepackage{xcolor}

\usepackage{lastpage}

\firstpageno{1}

\begin{document}

\title{\raisebox{-0.25\height}{\includegraphics[height=0.9em]
{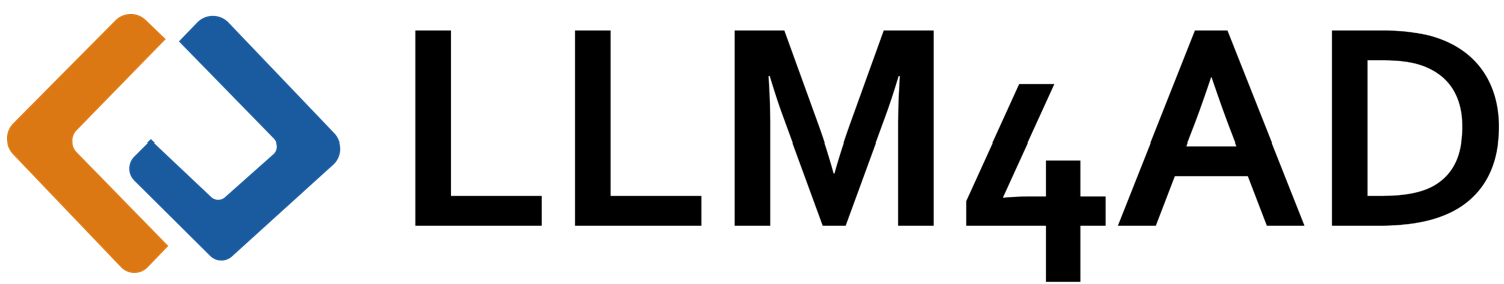}}: A Platform for Algorithm Design with Large Language Model}



\author{%
  \name Fei Liu$^{1,\star}$, 
  \name Rui Zhang$^{1,\star}$, 
  \name Zhuoliang Xie$^2$, 
  \name Rui Sun$^1$, 
  \name Kai Li$^2$, 
  \name Qinglong Hu$^1$, 
  \name Ping Guo$^1$, 
  \name Xi Lin$^3$, 
  \name Xialiang Tong$^4$, 
  \name Mingxuan Yuan$^4$, 
  \name Zhenkun Wang$^{2,\dagger}$, 
  \name Zhichao Lu$^{1,\dagger}$, 
  \name Qingfu Zhang$^{1,\dagger}$
  \AND
  \addr $^1$ City University of Hong Kong \\
  $^2$ Southern University of Science and Technology \\
  $^3$ Xi'an Jiaotong University \\
  $^4$ Huawei Technologies Co., Ltd.
}



{\def\thefootnote{$\star$}\footnotetext{Equal contribution.}}{}
{\def\thefootnote{\dag}\footnotetext{Corresponding authors.(wangzk3@sustech.edu.cn, zhichao.lu@cityu.edu.hk, qingfu.zhang@cityu.edu.hk)}}{}

\def\ourmethod{\texttt{LLM4AD}}

\maketitle

\vspace{10pt}

\begin{center} 
\href{https://github.com/Optima-CityU/LLM4AD}{\raisebox{-0.1\height}{\includegraphics[height=1em]
{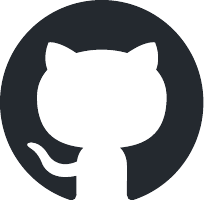}}
\textcolor{logocolor}{GitHub Repository and Examples}}
\end{center}

\vspace{10pt}

\begin{abstract}

We introduce \ourmethod{}, a unified Python platform for algorithm design (AD) with large language models (LLMs). 
\ourmethod{} is a generic framework with modularized blocks for search methods, algorithm design tasks, and LLM interface. 
The platform integrates numerous key methods and supports a wide range of algorithm design tasks across various domains including optimization, machine learning, and scientific discovery. 
We have also designed a unified evaluation sandbox to ensure a secure and robust assessment of algorithms. Additionally, we have compiled a comprehensive suite of support resources, including tutorials, examples, a user manual, online resources, and a dedicated graphical user interface (GUI) to enhance the usage of \ourmethod{}. 
We believe this platform will serve as a valuable tool for fostering future development in the merging research direction of LLM-assisted algorithm design. 
\end{abstract}

\vspace{10pt}

\section{Introduction}
Algorithms are pivotal in solving diverse problems across various fields such as industry, economics, healthcare, and technology~\citep{kleinberg2006algorithm,cormen2022introduction}. Traditionally, algorithm design has been a labor-intensive process requiring deep expertise. In the last three years, the use of large language models for algorithm design (\ourmethod{}) has emerged as a promising research area with the potential to fundamentally transform how algorithms are designed, optimized, and implemented~\citep{liu2024systematic}. The remarkable capabilities and flexibility of LLMs have shown potential in enhancing the algorithm design process, including performance prediction~\citep{hao2024large}, heuristic generation~\citep{liu2024evolution}, code optimization~\citep{hemberg2024evolving}, and even the creation of new algorithmic concepts~\citep{girotra2023ideas}. This approach not only reduces the human effort required in the design phase but also boosts the creativity and efficiency of the solutions produced~\citep{liu2024evolution,romera2024mathematical}.

Despite the rapid emergence of \ourmethod{} methods~\citep{liu2024systematic} and the expanding range of application domains~\citep{romera2024mathematical,liu2024evolution,ye2024reevo,yao2024evolve,guo2024autoda,guo2024two}, this area faces three challenges:
\begin{enumerate}
\item \textbf{Lack of an easy-to-use toolkit and platform.} It is difficult for researchers from diverse backgrounds to conduct studies at this intersection, as they need to address issues from both LLMs and algorithm design and implement their own code. 

\item \textbf{Lack of a generic implementation pipeline.} Existing works are implemented using different pipelines and programming languages. There are inconsistencies such as varied running times and LLMs queries. 

\item \textbf{Lack of benchmarks specifically for LLM-assisted algorithm design.} Existing works either compare selected instances or propose their own algorithm design tasks. Even when compared on the same tasks, different prompt engineering or templates can be used, leading to varied performance. A unified benchmark with a proper template and default settings would enable fair comparisons.
\end{enumerate}

This paper introduces \ourmethod{}, a unified Python library for LLM-based algorithm design that addresses these gaps. The platform integrates numerous key methods and supports a wide range of algorithm design tasks across various domains, including optimization, machine learning, and scientific discovery. We have also designed a unified evaluation sandbox to ensure a secure and robust assessment of algorithms. Additionally, we have compiled a comprehensive suite of support resources, including tutorials, examples, a user manual, online resources, and a graphical user interface (GUI) to enhance the usability of \ourmethod{}. We believe this platform will serve as a valuable tool by fostering usage and comparison in the emerging research direction on LLM-based algorithm design. The code is available at:
\url{https://github.com/Optima-CityU/LLM4AD}.

\section{\ourmethod{}}

\subsection{Framework}

\begin{figure}
    \centering
    \includegraphics[width=0.98\linewidth]{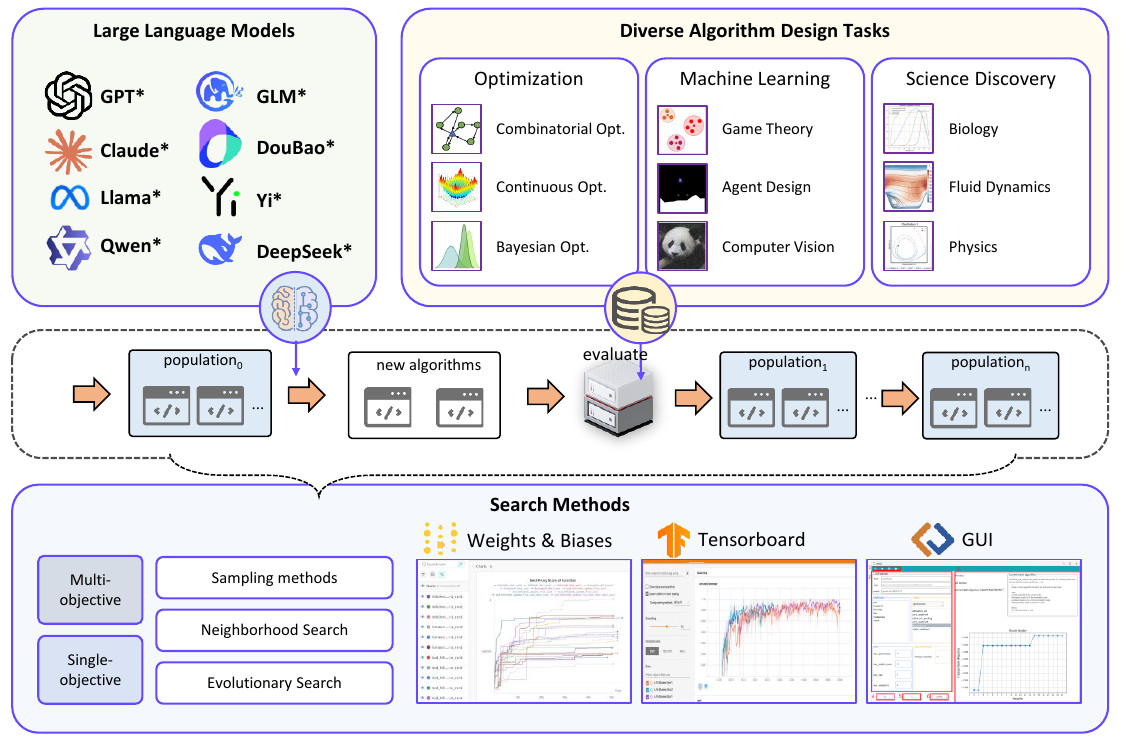}
    \caption{\ourmethod{} platform overview.}
    \label{fig:overview}
\end{figure}

As illustrated in Figure~\ref{fig:overview}, the platform consists of three blocks: 1) Search methods, 2) LLM interface, and 3) Task evaluation interface.

\begin{itemize}
    \item \textbf{Search methods:}
We build the pipeline with an iterative search framework, in which a population is maintained and elite algorithms are survived. 
\begin{itemize}
    \item Multiple Objectives: The task of designing algorithms may involve one or more objectives, such as optimizing performance and efficiency. Our approach incorporates both single-objective and multi-objective search methods. 
    \item Population Size: In many search methods, e.g., neighbourhood search methods, the population size can be set to one. 
\end{itemize}

\item \textbf{LLM interface:}
LLMs are guided by prompts to create new algorithms.  These LLMs can be utilized either locally or remotely, with a unified interface available for both options. To enhance efficiency, we have implemented parallelization in the sampling process.
\item \textbf{Tasks evaluation interface:}
In each generation, algorithms for a specific problem are assessed on a set of instances on the target algorithm design task and given a fitness score. Our evaluation sandbox guarantees a secure, effective, and manageable evaluation process for various problems and implementations.
\end{itemize}

\subsection{Usage}
\subsubsection{Script Usage }
\begin{itemize}
    \item \textbf{Import} method, tasks, and llm interface. 
    \item \textbf{LLM} interface initialization: Set up host endpoint, key and llm.
    \item \textbf{Task evaluation} interface initialization: Initialize the evaluation interface with settings to manage the evaluation process, e.g., maximum wall-clock time for each evaluation.
    \item \textbf{Search method} initialization: Set up the Profiler and parameters. Pass LLM and Task interfaces into the search procedure.
    \item \textbf{Run}: Run the search process. Logs will be recorded and displayed according to the Profiler settings.
\end{itemize}
One example script is as follows.


    





\begin{lstlisting}
from llm4ad.task.optimization.online_bin_packing import OBPEvaluation
from llm4ad.tools.llm.llm_api_https import HttpsApi
from llm4ad.method.eoh import EoH, EoHProfiler

if __name__ == "__main__":
    llm = HttpsApi(
        host="xxx",   # your host endpoint, e.g., api.openai.com
        key="sk-xxx", # your key, e.g., sk-xxxxxxxxxx
        model="xxx",  # your llm, e.g., gpt-3.5-turbo
        timeout=20
    )
    method = EoH(
        llm=llm,
        profiler=EoHProfiler(log_dir="logs/eoh", log_style="simple"),
        evaluation=OBPEvaluation(),
        max_generations=10,
        pop_size=4,
        num_samplers=4,
        num_evaluators=4,
    )
    method.run()
\end{lstlisting}

\subsubsection{GUI Usage}\label{sec:appx_b_gui}



    




\ourmethod{} provides an easy-to-use graphical user interface (GUI). Through this GUI, users can easily configure settings, execute experiments, and monitor results without any coding knowledge. This interface simplifies user interaction, making the \ourmethod{} platform more accessible and easier to use.

\begin{figure}[htbp]
    \centering
    \includegraphics[width=0.98\linewidth]{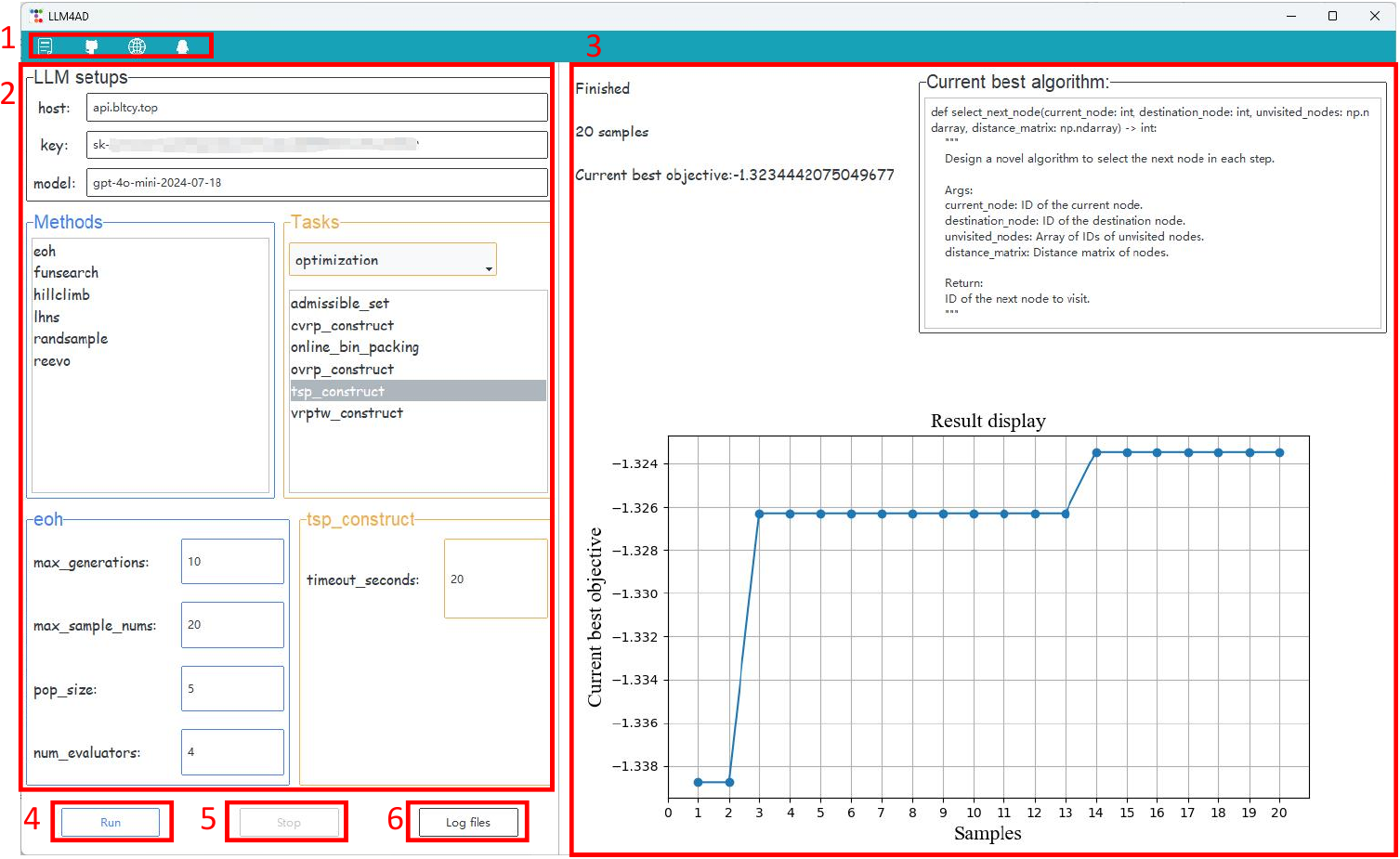}
    \caption{Graphical user interface (GUI) for \ourmethod{}.}
    \label{fig:gui_image}
\end{figure}

The GUI is launched by executing the \textit{run\_gui.py} Python script. 
As shown in Figure \ref{fig:gui_image}, the main window of GUI includes six components: 1): \textit{Menu bar}; 2): \textit{Configuration panel}; 3): \textit{Results dashboard}; 4): \textit{Run} button; 5): \textit{Stop} button; 6): \textit{Log files} button.

The \textit{Menu bar} offers quick access to various resources, such as documentation or the website of the \ourmethod{} platform, through clickable buttons that redirect users to the relevant pages.
To conduct experiments via the GUI, users should
\begin{itemize}[itemsep=-2pt]
    \item Set up LLM interface. Set up the parameters of the LLM interface in the \textit{Configuration panel}.
    These parameters include the internet protocol (IP) address of the application programming interface (API) provider, an API key, and the name of the LLM. 
    \item Set up Search method and Algorithm design task. Users can also select the search method and the algorithm design task by clicking. 
For the chosen method and task, specific parameters such as \textit{max\_samples} (the maximum number of LLM invocations) can be configured.
\end{itemize}
After setting all configurations, the experiment can be started by clicking the \textit{Run} button.
The \textit{Results dashboard} then displays the experimental results such as the convergence curve of the objective values and the currently best-performing algorithm along with its corresponding objective value.
During an experiment, users can stop the process using the \textit{Stop} button or access detailed experimental results through the \textit{Log files} button.

The current version of GUI only supports conducting experiments with a single method under a single LLM configuration each time. In the future, we plan to extend the GUI to enable batch experiments.

\subsection{Search Methods}

Search methods are crucial for effective LLM-based algorithm design. Recent studies have shown that standalone LLMs, even when enhanced with various prompt engineering techniques, are often insufficient for many algorithm design tasks~\citep{zhang2024understanding}. We have integrated a variety of search methods, including simple sampling, commonly used single-objective evolutionary search methods, multi-objective evolutionary search, and various neighborhood searches.

\begin{itemize}
 \item {Single-objective Search}
 \begin{itemize}
    \item Sampling: repeated sampling.
    \item Neighborhood search: tabu search, iterated local search, simulated annealing, variable neighbourhood search.
    \item Evolutionary search: EoH~\citep{liu2024evolution}, FunSearch~\citep{romera2024mathematical}, (1+1)-EPS~\citep{zhang2024understanding}
 \end{itemize}

 \item {Multi-objective Search}
 \begin{itemize}
    \item Multi-objective evolutionary search: MEoH~\citep{yao2024multi}, NSGA-II~\citep{deb2002fast}, MOEA/D~\citep{zhang2007moea}
 \end{itemize}

\end{itemize}

An abstract base method is provided to modularize the essential format and functions of these methods, maintaining flexibility to facilitate easy extension and implementation of custom search methods by users. 

Each method is equipped with three profilers: 1) base profiler, 2) Tensorboard profiler, and 3) Weights \& Biases (wandb) profiler, to meet diverse user requirements.

\subsection{Evaluation Interface and Tasks}

\subsubsection{Tasks}

As illustrated in Figure~\ref{fig:overview}, \ourmethod{} is applicable to a broad range of algorithm design domains including \\
\begin{itemize}[itemsep=-15pt]
    \item \textbf{Optimization}: combinatorial optimization~\citep{liu2024evolution,ye2024reevo}, continuous optimization, surrogate-based optimization~\citep{yao2024evolve}. \\
    \item \textbf{Machine learning}: agent design~\citep{hu2024automated}, computer vision~\citep{guo2024autoda}. \\
    \item \textbf{Science discovery}: biology~\citep{shojaee2024llmsr}, chemistry, physics, fluid dynamics~\citep{zhang2024autoturb} and Feynman Equation~\citep{matsubara2022srsd}. \\
    \item \textbf{Others}: game theory, mathematics~\citep{romera2024mathematical}, etc.
\end{itemize}

As illustrated in Table~\ref{tab:tasks}, the platform includes a diverse collection of over 20 tasks (there will be 160+ tasks soon) from various domains such as optimization, machine learning, and scientific discovery. These tasks are quick to evaluate and have clearly defined formulations for easy comparison.

\begin{table}[th]
    \caption{Algorithm design tasks in \ourmethod{}.}
    \centering
    \large
    \resizebox{1\textwidth}{!}{
    \begin{threeparttable}
        \begin{tabular}{c|c}
        \toprule
        Task Type & Task Name (Abbreviation)\\
        \midrule
        \multirow{8}{*}{Optimization} & Capacitated Vehicle Routing Problem (CVRP, 2 tasks), \\
        & Open Vehicle Routing Problem (OVRP, 2 tasks) \\
        & Online Bin Packing (OBP, 1 task), Traveling Salesman Problem (TSP, 2 tasks), \\ 
        & Vehicle Routing Problem with Time Window (VRPTW, 2 tasks), \\
        & Admissible Sets (SET, 1 task), Flow Shop Scheduling Problem* (FSSP, 2 tasks), \\ 
        & Evolution Algorithm* (EA, 1 task), Multi-objective evolutionary Algorithm (MEA, 1 task), \\
        & Maximum Cut Problem* (MCP, 1 task), Knapsack Problem* (MKP, 1 task), \\
        & Surrogate-based optimization (1 task)\\
        \midrule
        \multirow{3}{*}{Machine Learning} 
        & Acrobot (ACRO, 1 task), Mountain Car (CAR, 1 task), \\
        & Moon Lander (ML, 1 task), Cart Pole (CARP, 1 task), \\
        & Mountain Car Continuous* (CARC, 1 task), Pendulum* (PEN, 1 task), \\
        \midrule
        \multirow{4}{*}{Scientific Discovery} & Bacterial Growth (BACT, 1 task), Nonlinear Oscillators (OSC, 2 tasks), \\
        & Material Stress Behavior (MSB, 1 task), Ordinary Differential Equations (ODE, 16 tasks), \\
        & SRSD-Feynman Easy Set* (SRSD-E, 30 tasks), SRSD-Feynman Medium Set* (SRSD-M, 40 tasks), \\
        & SRSD-Feynman Hard Set* (SRSD-H, 50 tasks) \\
        \bottomrule
    \end{tabular}
        \begin{tablenotes}
        \item There are \textbf{160+} tasks added or being added (marked with *).
        \end{tablenotes}
    \end{threeparttable}  
    }
    \label{tab:tasks}
\end{table}

\subsubsection{Examples}

We also offer a variety set of example algorithm design tasks. These examples are used for 1) demonstrating different settings and 2) showcasing more complex tasks on local algorithm design tasks.

\subsubsection{Evaluation Sandbox}

A secure evaluation sandbox is provided, enabling the safe and configurable evaluation of generated code. This includes optional optimizations and safety features such as timeout handling and protected division.

\subsection{LLM Interface}
We have provided a general LLM interface tailored for iterative algorithm search. This interface supports two types of demo interactions:
\begin{itemize}
    \item Remote API interaction: We offer OPENAI format API interaction, suitable for most LLM API requests.
    \item Local deployment: A guide and template for local open-source LLM deployment are provided.
\end{itemize}
Both interfaces are modularized to ensure efficiency and control, with features including parallel processing, time control, and failure detection.

\section{Benchmark Results}

\subsection{Settings}

We choose four search methods in our platform with consistent benchmark settings. We initialize all compared methods with the respective template algorithm on each problem. Table \ref{tab:bench_setting} summarizes the benchmark hyper-parameter settings.

\begin{table}[ht]
\caption{Summary of benchmark settings.}
\centering
\resizebox{0.75\textwidth}{!}{%
\label{tab:bench_setting}
\begin{tabular}{@{\hspace{2mm}}l|@{\hspace{2mm}}c}
\toprule
Description of Setting & Value \\ \midrule
Maximum number of function evaluations (\#FE) & 2,000 \\ 
Population size (for EoH) & 10 \\ 
\# of islands, \# of samples per prompt (for FunSearch) & 10, 4 \\
Number of independent runs per experiment & 3 \\ \midrule 
\begin{tabular}[c]{@{}l@{}}Maximum evaluation time for each algorithm\\(to cope with invalid algorithms, such as infinite loops) \end{tabular} & \begin{tabular}[c]{@{}l@{}} 50 seconds\end{tabular}\\
\bottomrule
\end{tabular}
}
\end{table}

We investigate a subset of nine algorithm design tasks provided by our platform, encompassing machine learning, combinatorial optimization, and scientific discovery scenarios.
The included tasks are summarized in Table \ref{tab:test_tasks}. 

\begin{table}[h]
    \caption{Tested algorithm design tasks.}
    \centering
    \resizebox{0.8\linewidth}{!}{
        \begin{tabular}{c|c}
        \toprule
        Task Type & Task Name (Abbreviation)\\
        \midrule
        \makecell{Machine\\Learning} & 
        \makecell{Acrobot (ACRO), Mountain Car (CAR)} \\
        \midrule
        \makecell{Combinatorial\\Optimization} & \makecell{Cpacitated Vehicle Routing Problem (CVRP), \\Online Bin Packing (OBP), Traveling Salesman Problem (TSP), \\Vehicle Routing Problem with Time Window (VRPTW)}\\
        \midrule
        \makecell{Scientific\\Discovery} & \makecell{Bacterial Growth (BACT), Admissible Sets (SET),\\Nonlinear Oscillators (OSC)} \\
        \bottomrule
    \end{tabular}
    }

    \label{tab:test_tasks}
\end{table}

We use a diverse set of eight open-source and closed-source general-purposed LLMs, i.e., Llama-3.1-8B,
Yi-34b-Chat, GLM-3-Turbo, Claude-3-Haiku, Doubao-pro-4k, GPT-3.5-Turbo, GPT-4o-Mini, and Qwen-Turbo, and compare their results on nine automated algorithm design tasks.

\begin{table}[h]
    \caption{Overview of the LLMs evaluated in the experiment. We use performance on ``HumanEval'' and ``MMLU'' to indicate the capabilities of LLMs on code and general knowledge.}
    \centering
    \resizebox{0.8\linewidth}{!}{
    \begin{tabular}{lccc}
    \toprule
    \textbf{Model} & \textbf{Open Source} & \textbf{HumanEval} & \textbf{MMLU} \\
    \midrule
    Llama-3.1-8B~\citep{dubey2024llama} & \checkmark & 72.6 & 69.4\\
    Yi-34b-Chat~\citep{young2024yi} & \checkmark & 75.2 & 76.8 \\
    GLM-3-Turbo~\citep{glm2024chatglm} & \checkmark & 70.1 & 74.3 \\
    \midrule
    Claude-3-Haiku~\citep{anthropic2024Claude3} & \ding{53} & 75.9 & 75.2 \\
    Doubao-pro-4k & \ding{53} & 73.0 & 78.0 \\
    GPT-3.5-Turbo~\citep{ye2023comprehensive} & \ding{53} & 60.3 & 70.0 \\
    GPT-4o-Mini~\citep{achiam2023gpt} & \ding{53} & 87.2 & 82.0 \\
    Qwen-Turbo~\citep{yang2024qwen2} & \ding{53} & 86.6 & 86.1 \\
    \bottomrule
    \end{tabular}
    }
    \label{tab:llm}
\end{table}



\begin{figure}[t]
     \centering
     \begin{subfigure}[b]{0.98\textwidth}
         \centering
         \includegraphics[width=\textwidth]{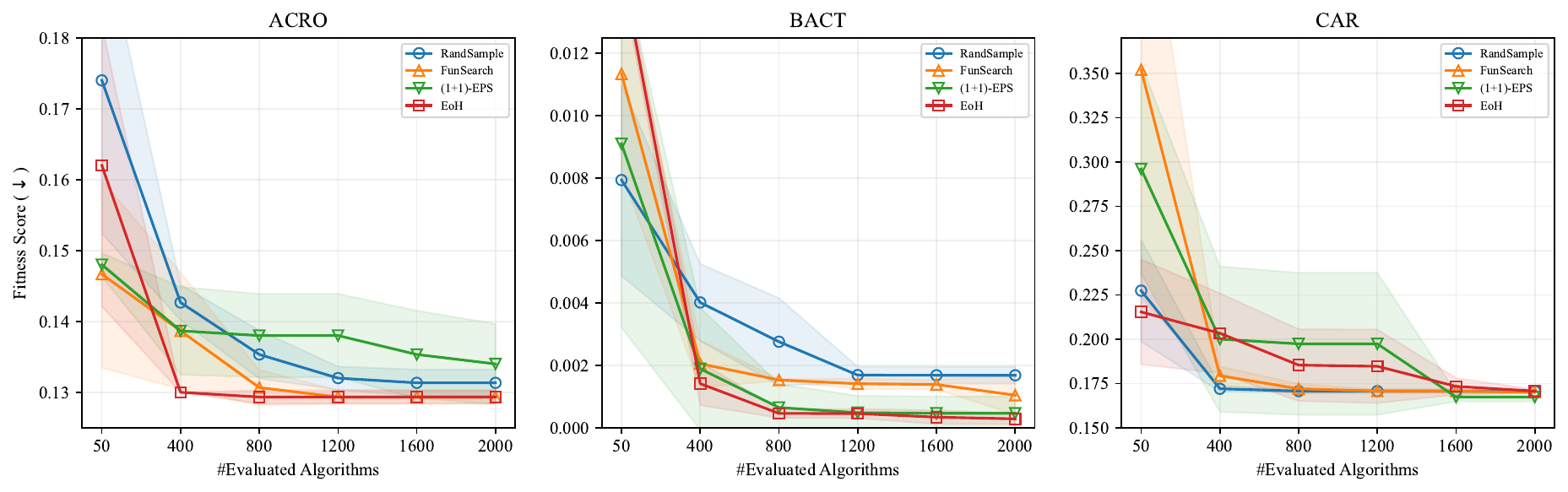}
     \end{subfigure}\\
     \begin{subfigure}[b]{0.98\textwidth}
         \centering
         \includegraphics[width=\textwidth]{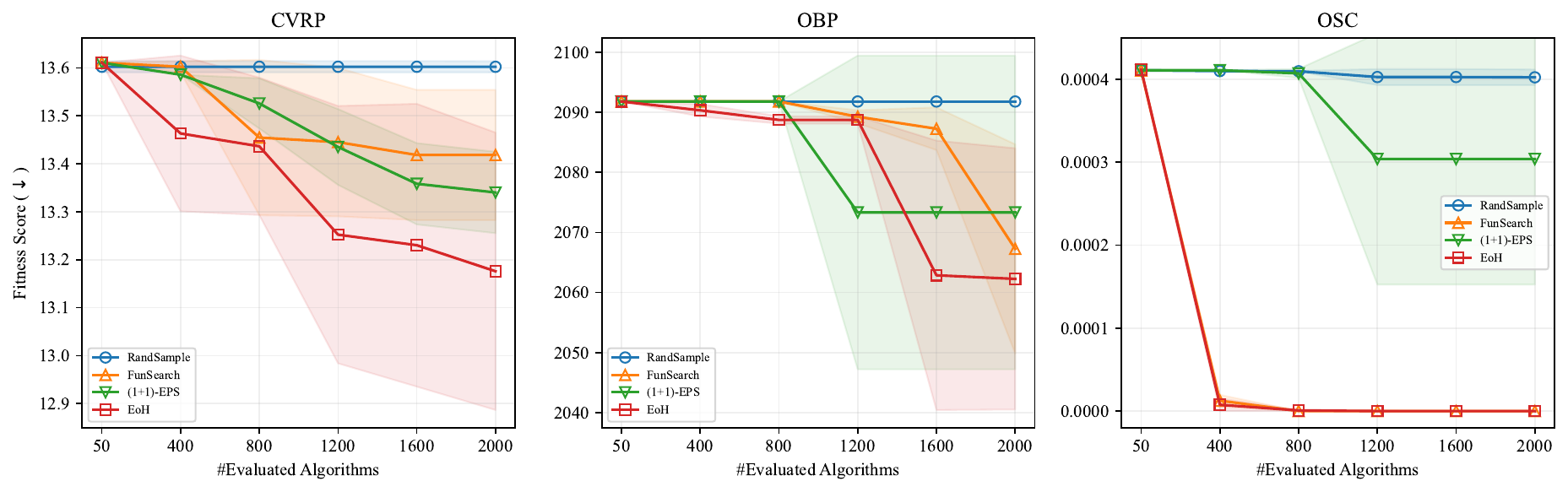}
     \end{subfigure}\\
          \begin{subfigure}[b]{0.98\textwidth}
         \centering
         \includegraphics[width=\textwidth]{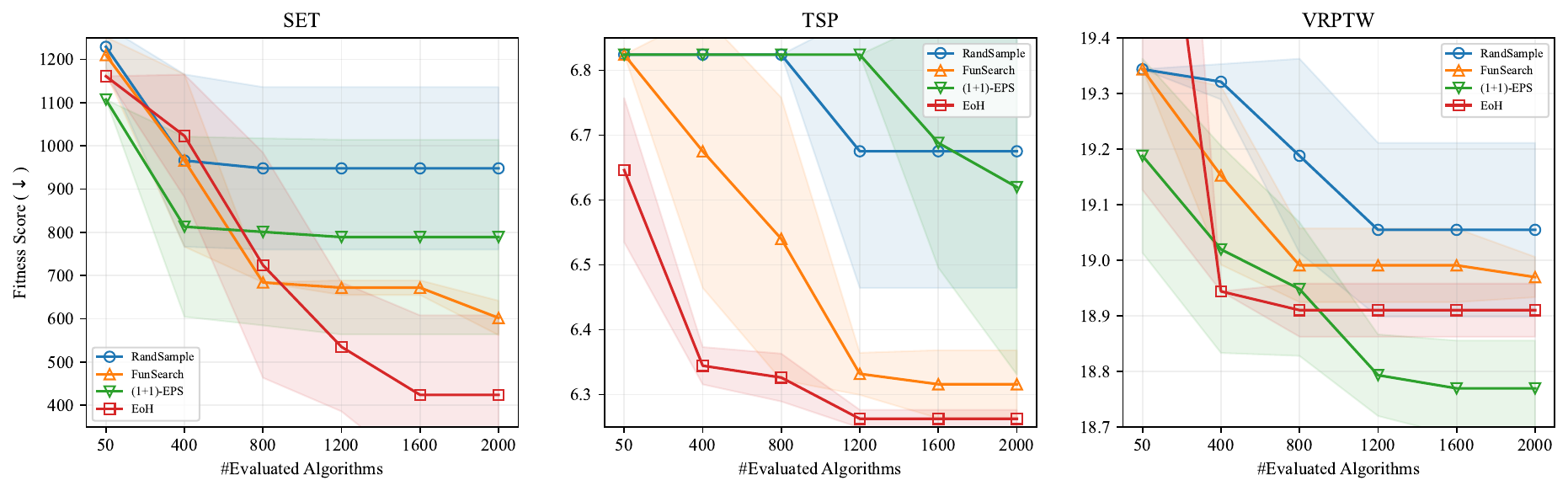}
     \end{subfigure}
     
    \caption{Convergence curve comparison on the performance (measured by fitness score) of the top-1 algorithms generated by GPT-4o-Mini. The performance averaged over three independent runs are denoted with markers (lower the better), while the standard deviations of scores are highlighted with the shaded regions.}
    \label{fig:converge}
\end{figure}

\subsection{Results on Different Tasks} Fig.~\ref{fig:converge} demonstrates the convergence curve of the performance of the top-1 algorithms generated by GPT-4o-Mini. 
The performance is measured by the objective score on each task.
The mean and standard deviation performance over three independent runs are denoted as markers and shaded areas, respectively. 
We draw observations from the experimental results that:
\begin{itemize}
    \item \ourmethod{} is applicable to diverse algorithm design tasks and application scenarios.
    \item Methods coupled with a search strategy, i.e., EoH, (1+1)-EPS, and FunSearch, significantly outperform random sampling on most tasks. This underscores the importance of synergizing LLMs with search. 
    \item All tested methods demonstrate marginal performance variations on the Mountain Car problem. However, there is a noticeable performance gap between different methods on OSC, SET, TSP, and VRPTW problems. This reveals that tasks provided in our platform have different difficulties.
    \item EoH and FunSearch exhibit better performance on most tasks due to their inherent diversity control mechanism. While the performance of (1+1)-EPS variants significantly across different tasks due to its greedy nature.
\end{itemize}

\begin{figure}[t]
    \centering
    \includegraphics[width=0.98\linewidth]{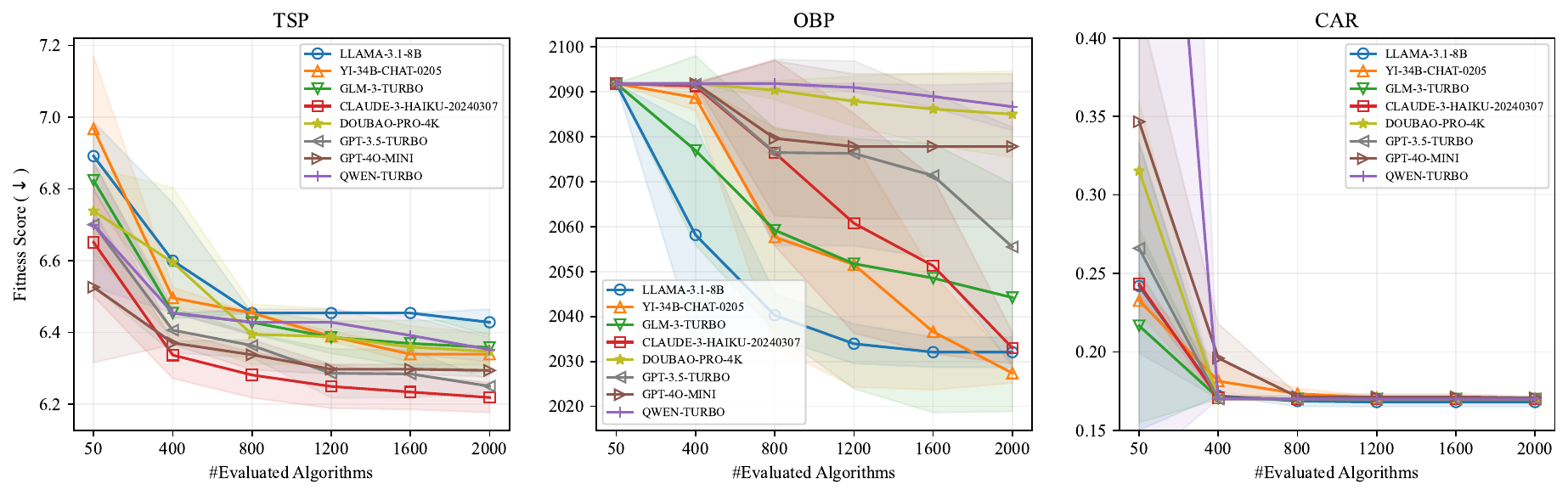}
    \caption{Convergence curve comparison on the performance of the top-1 heuristics generated by various LLMs. The mean score aggregated over three independent runs are denoted with markers (lower the better), while the standard deviations of scores are highlighted with the shaded regions. 
    \label{fig:llms_converge}}
\end{figure}

\begin{figure}[t]
    \centering
    \includegraphics[width=1\linewidth]{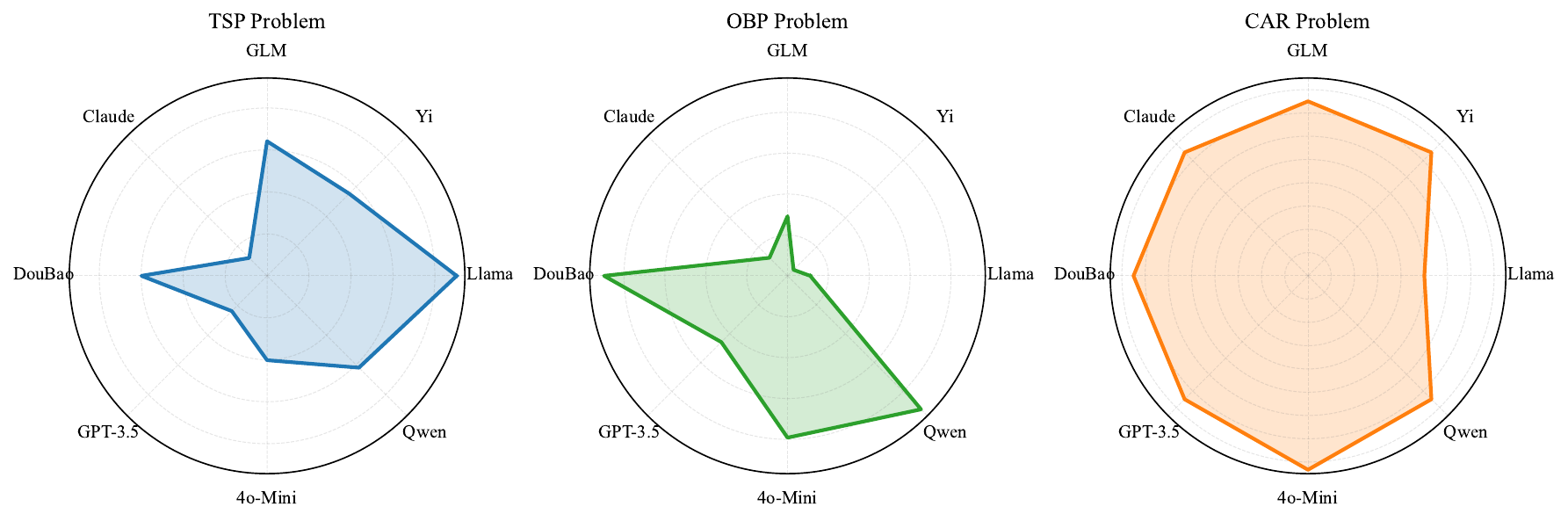}
    \caption{Radar plot on the performance of the top-1 algorithm generated by the EoH method using different LLMs. The radius of each vertex is calculated by the normalized fitness value over three independent runs; hence, a smaller radius/enclosed area indicates better performance.}
    \label{fig:llms_radar}
\end{figure}

\subsection{Results with Different LLMs}
Fig.~\ref{fig:llms_converge} and Fog.~\ref{fig:llms_radar} compare the performance of various LLMs on three tasks. We can summarize from the results that:

\begin{itemize}
    \item \ourmethod{} can easily integrate open- and close-source LLMs with different capabilities.
    \item There are significant variances in performance attributable to the choice of LLM on three tasks, with the notable exception of the CAR problem where this variance is marginal.
    \item LLMs with more coding capability (denoted by a better HumanEval score as demonstrated in Table \ref{tab:llm}) do not necessarily lead to better performance on AD problems. 
    \item No single LLM exhibits notable superiority over others, as the standard deviations are overlapped in Fig.~\ref{fig:llms_converge}. This indicates that our platform supports a diverse set of LLMs.
\end{itemize}

\section{Extensibility}

\subsection{Add New Methods}
\ourmethod{} platform encourages users to perform further optimization and customization to existing algorithm design methods. Our support for developers can be summarized in two perspectives as follows:

\begin{itemize}
    \item \ourmethod{} has fully open-sourced code implementations for various evolutionary search-based algorithm design methods for reference. The source code incorporates implementations of various population management strategies such as genetic algorithm (implemented in EoH), island model (implemented in FunSearch), and 1+1 search (implemented in (1+1)-EPS)). In addition, \ourmethod{} has integrated diverse test cases (algorithm design tasks) and LLMs with unified interfaces, enabling prompt validation and debugging during development. 
    \item \ourmethod{} provides useful tools and APIs to help code manipulation, secure evaluation, and task definition. Besides, \ourmethod{} has released elaborated documents\footnote{https://llm4ad-doc.readthedocs.io/en/latest/} as well as \emph{jupyter-notebooks} for each module, aiming to demonstrate and visualize the functionality and effect of each module and each API. We believe this will foster in-depth comprehension of our search and code-manipulating pipeline. 
\end{itemize}

\subsection{Add New Tasks}
\ourmethod{} is designed to be easily applicable to customized algorithm design tasks and has unified the evaluation interface for each algorithm design task. There are two major steps to apply \ourmethod{} to a specified algorithm design task:
\begin{itemize}
    \item Extend the \emph{llm4ad.base.Evaluation} interface class and override the \emph{evaluate\_program()} method, which defines how to measure the objective score of a searched algorithm. Users can also restrict the timeout seconds during evaluation and perform \emph{numba.jit} acceleration through setting the corresponding arguments. 
    \item Specify an executable template program that comprises Python packages import, a function call with input/output types, a doc-string with the exact meaning of each argument, and a function body to show an example implementation. Since the template program will be assembled to the prompt content in the search process later, providing informative and precise doc-string is therefore required.
\end{itemize}
Once the modified evaluation instance is passed into the search method, \ourmethod{} will automatically invoke the specified evaluation method and perform a secure evaluation (prevent algorithms that may be harmful to the search pipeline, e.g., abort the search or endless loop) of each algorithm code.

\subsection{Add New LLM Sampler}
A sampler defines and specifies the method to access an LLM. For instance, users can choose to either query remote LLMs (e.g., GPT-4o) using HTTPS requests, or infer a locally deployed LLM using inference libraries (e.g., \emph{transformers}, and \emph{vLLM}). To increase the extensibility of samplers, \ourmethod{} defines an interface \emph{llm4ad.base.Sampler} where the \emph{draw\_sample()} function leaves unimplemented. Users are able to customize their sampler by overriding the method and passing the user-defined-sampler instance to a search method.

\section{Conclusion}
In conclusion, \ourmethod{} stands as a comprehensive and unified Python platform tailored for the design of algorithms using large language models. It features a generic framework with modularized components, including search methods, algorithm design tasks, and an LLM interface, catering to a broad spectrum of domains such as optimization, machine learning, and scientific discovery. The platform is enriched with a robust evaluation sandbox to ensure secure and reliable algorithm assessment, alongside a wealth of support resources like tutorials, examples, a user manual, online resources, and a dedicated GUI. These elements collectively enhance the user experience and utility of \ourmethod{}. We are confident that \ourmethod{} will significantly contribute to the advancement and standardization of LLM-based algorithm design, promoting extensive usage and facilitating comparative research in this emerging field. Through these efforts, \ourmethod{} aims to accelerate innovation and exploration in algorithm design, leveraging the capabilities of large language models.


\section{Acknowledgment}
Thank Zhiling Mao, Shunyu Yao, Yiming Yao, Ping Guo, and Zhiyuan Yang for their suggestion and help in the development of \ourmethod{}. 

\vskip 0.2in
\bibliography{jmlr}


\end{document}